\documentclass[graybox]{svmult}

\usepackage{mathptmx}       % selects Times Roman as basic font
\usepackage{helvet}         % selects Helvetica as sans-serif font
\usepackage{courier}        % selects Courier as typewriter font
\usepackage{type1cm}        % activate if the above 3 fonts are
\usepackage{makeidx}         % allows index generation
\usepackage{graphicx}        % standard LaTeX graphics tool
\usepackage{multicol}        % used for the two-column index
\usepackage[bottom]{footmisc}% places footnotes at page bottom

\usepackage{placeins}

\makeindex             % used for the subject index
\begin{document}

\title*{Analysis of a chaotic spiking neural model: The NDS neuron}
\author{Mohammad Alhawarat, Waleed Nazih and Mohammad Eldesouki}
\institute{Mohammad Alhawarat \at Department of Computer Science,
                                  College of Computer Engineering and Sciences, \\
                                  Salman Bin Abdulaziz University, Kingdom of Saudi Arabia,\\
                                  Tel.: +966-1-1588371, Fax: +966-1-1588302, \email{m.alhawarat@sau.edu.sa}
\and Waleed Nazih \at             Department of Computer Science,
                                  College of Computer Engineering and Sciences, \\
                                  Salman Bin Abdulaziz University, Kingdom of Saudi Arabia,\\
                                  Tel.: +966-1-1588371, Fax: +966-1-1588302, \email{w.nazeeh@sau.edu.sa}
\and Mohammad Eldesouki \at       Department of Information Systems,
                                  College of Computer Engineering and Sciences, \\
                                  Salman Bin Abdulaziz University, Kingdom of Saudi Arabia,\\
                                  Tel.: +966-1-1588371, Fax: +966-1-1588302, \email{m.eldesouki@sau.edu.sa}}

%
%\keywords{Nonlinear dynamics, R\"{o}ssler attractor, Chaos, Discretization}

\maketitle

\abstract*{Further analysis and experimentation is carried out in this paper for a chaotic dynamic model, viz. the Nonlinear Dynamic State neuron (NDS). The analysis and experimentations are performed to further understand the underlying dynamics of the model and enhance it as well. Chaos provides many interesting properties that can be exploited to achieve computational tasks. Such properties are sensitivity to initial conditions, space filling, control and synchronization.Chaos might play an important role in information processing tasks in human brain as suggested by biologists. If artificial neural networks (ANNs) is equipped with chaos then it will enrich the dynamic behaviours of such networks.
The NDS model has some limitations and can be overcome in different ways. In this paper different approaches are followed to push the boundaries of the NDS model in order to enhance it. One way is to study the effects of scaling the parameters of the chaotic equations of the NDS model and study the resulted dynamics. Another way is to study the method that is used in discretization of the original R\"{o}ssler that the NDS model is based on. These approaches have revealed some facts about the NDS attractor and suggest why such a model can be stabilized to large number of unstable periodic orbits (UPOs) which might correspond to memories in phase space.}

\abstract{Further analysis and experimentation is carried out in this paper for a chaotic dynamic model, viz. the Nonlinear Dynamic State neuron (NDS). The analysis and experimentations are performed to further understand the underlying dynamics of the model and enhance it as well. Chaos provides many interesting properties that can be exploited to achieve computational tasks. Such properties are sensitivity to initial conditions, space filling, control and synchronization.Chaos might play an important role in information processing tasks in human brain as suggested by biologists. If artificial neural networks (ANNs) is equipped with chaos then it will enrich the dynamic behaviours of such networks.
The NDS model has some limitations and can be overcome in different ways. In this paper different approaches are followed to push the boundaries of the NDS model in order to enhance it. One way is to study the effects of scaling the parameters of the chaotic equations of the NDS model and study the resulted dynamics. Another way is to study the method that is used in discretization of the original R\"{o}ssler that the NDS model is based on. These approaches have revealed some facts about the NDS attractor and suggest why such a model can be stabilized to large number of unstable periodic orbits (UPOs) which might correspond to memories in phase space.}

\section{Introduction}
\label{sec:intro}

Chaos might play an important role in information processing tasks in human brain as shown in \cite{Babloyantz1996,Destexhe1994,Freeman1999,Freeman2000,Freeman1994b,Rapp1993,Theiler1995,Wu2009}.
Some properties that might be useful for information processing tasks are: sensitivity to initial conditions, space filling, control, synchronization and a rich dynamics that can be accessed using different control methods. In theory, if Artificial Neural Networks (ANN) are equipped with chaos they will enable a large number of rich dynamic behaviors. After applying control, these dynamics can be accessed using one of the control mechanisms such as feedback control \cite{Ott1990,Pasemann1998,Pyragas1992}. Applying such control mechanisms to discrete chaotic neural models showed that the model would stabilize into one of many UPOs that are embedded in the chaotic attractor.

Different chaotic neural models have been devised in recent years to explore the possibilities of exploiting the rich dynamics that such models might provide for information processing tasks. One of these model is the NDS model \cite{Crook2005}. This model is based on R\"{o}ssler system \cite{Rossler1976}.

R\"{o}ssler is a simple chaotic system. It has been studied many times in terms of control investigation and biological studies \cite{YDing2010,Bershadskii2011} to name a few.

The origins of this model dates back to 2003 where the authors in\cite{Crook2003} have proposed a chaotic neuron that is based on R\"{o}ssler system. The idea was to exploit the rich dynamics of the chaotic attractor to represent internal states and therefore the chaotic attractor can represent an infinite state machine. Many experiments have been carried out to show that using periodic input signals would cause the chaotic attractor to stabilize to an UPO. The control mechanism used was a modified version of Pyragas\cite{Pyragas1992} where the period length τ is considered a system variable. Small networks of $2-3$ neurons have been studied and the network has stabilized to one UPO according to a periodic length that is implicitly appears in the input pattern.

The model is very interesting due to the fact that it theoretically allows an access to a large number of UPOs, which correspond to memories in phase space, using only single NDS neuron. In contrast, the Hopfield neural network can give only 0.15n memory size (where n is the number of neurons in the network).

The NDS model is studied in a series of works \cite{Crook2005,Goh2007,Alhawarat2007,Crook2008,Aoun2010}. The authors in\cite{Goh2007} have used Lorenz attractor instead of R\"{o}ssler. They have used transient computation machine to detect human motion from sequences of video frames. In another paper\cite{Crook2008} the authors argued that chaos may equip mammalian brain with the equivalent of kernel trick for solving hard nonlinear problems.

In \cite{Aoun2010} networks of NDS neurons have been investigated in the context of Spike Time Dependent Plasticity (STDP) which is a property of cortical neurons. The author has suggested that NDS neurons may own the realism of biological neural networks; this has been supported by experiments conducted by the author.

The NDS model has been investigated thoroughly in\cite{Alhawarat2007}. In his investigation, the author has studied the chaotic behavior of the model from both experimental and analytical perspectives. Explanation of the behavior of the model has become clear after the experimentations and the mathematical analysis and the study has shown interesting results.

In this paper some of the limitations that exist in the NDS model will be investigated. This includes tuning the model parameters for the sake of enhancing the model capacity in terms of the number of successfully stabilized UPOs. Moreover, the discretization method that used to to convert the continuous R\"{o}ssler system into the discrete NDS model will be discussed and compared to other well-known methods of discretization.

The paper is organized as follows: in section~\ref{sec:rossler} the original R\"{o}ssler model is introduced, in section~\ref{sec:nds_model} the NDS model is described, in section~\ref{sec:discrete} a mathematical analysis of the NDS model and related discretization methods are discussed, section~\ref{sec:tuning} is devoted to describe the experimentation setups that are carried out to tune the parameters of the NDS model, section~\ref{sec:discussion} includes discussions and finally section~\ref{sec:conclusion} concludes the paper.

\section{R\"{o}ssler Chaotic attractor}
\label{sec:rossler}
The R\"{o}ssler system \cite{Rossler1976} is a simple dynamical system that exhibits chaos and has only one nonlinear term in its equations. R\"{o}ssler built the system in 1976; it describes chemical fluctuation and is represented by the following differential equations:

\begin{equation}
\label{eqn:rslr_x} x' = -y-u
\end{equation}

\begin{equation}
\label{eqn:rslr_y} y' =x+a*y
\end{equation}

\begin{equation}
\label{eqn:rslr_z} z' =b+z(x-c)
\end{equation}

\begin{figure}[ht]
\centerline{\includegraphics[scale=0.75]{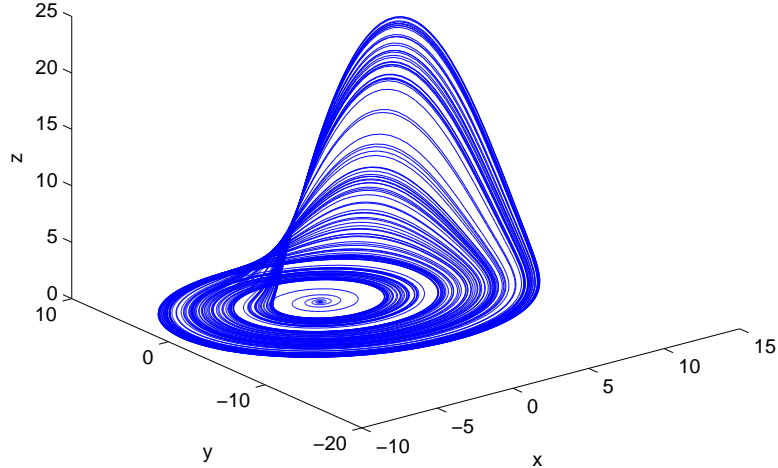}}
\caption{The R\"{o}ssler chaotic attractor with parameters $a= 0.2$, $b=0.2$, and $c=5.7$.}
\label{fig:FIG01}
\end{figure}

Where $a$ and $b$ are usually fixed and $c$ is varied and is called the control parameter. The familiar parameter settings for the R\"{o}ssler attractor are  $a= 0.2$, $b=0.2$, and $c=5.7$, and the corresponding attractor is shown in figure~\ref{fig:FIG01}. Note that the R\"{o}ssler attractor most of the time lies in the $x-y$ plane and comprises a disk that has a dense number of orbits. Note also that these orbits are stretching as a result of divergence and sensitivity to initial conditions. From time to time the R\"{o}ssler attractor rises in the $z$ direction and then folds back to the disk which forms a fin-like shape. The folding and stretching keep the R\"{o}ssler attractor bounded in phase space.

If a trajectory of a chaotic system evolved starting from an initial point within the attractor for a long period of time, then that trajectory will fill a bounded part of the phase space and the attractor of the system will have a fractional dimension. This bounded space is one of the properties of chaos, and is due to the attracting and the repelling of the trajectory by the fixed points that govern and organize the system behaviour. The type of these points determines the shape of the resulting attractor. This fractal dimension can be recognized in figure~\ref{fig:FIG01} where the attractor is not filling the whole space, instead it is filling part of the space.

\paragraph{Unstable periodic orbits}
An UPO is one of the dynamic behaviours that a nonlinear system exhibit in phase space. An UPO is a repeating orbit and is unstable as a result of being attracted and repelled by fixed points of an attractor. Chaotic attractors usually have a dense number of UPOs which can be accessed using controlling methods.

\begin{figure}[ht]
\centerline{\includegraphics[scale=0.75]{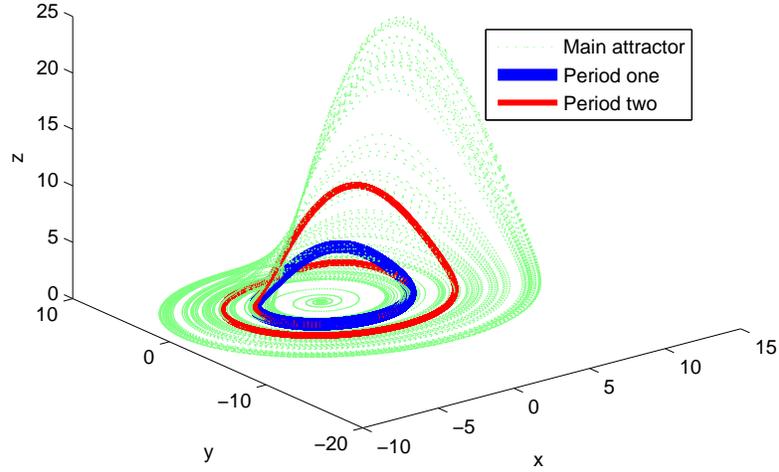}}
\caption{Two UPOs, one of period $1$ and the other of period $2$ in the R\"{o}ssler chaotic attractor occurred when parameters $a,b$ are fixed to $0.2$ and parameter $c$ is set to $2.5$ and $3.5$.}
\label{fig:FIG02}
\end{figure}

For example figure~\ref{fig:FIG02} shows two UPOs of period $1$ and period $2$ in the R\"{o}ssler attractor when the $c$ parameter is tuned to $2.5$ and $3.5$ respectively. Period $2$ here means that the UPO repeat twice in the attractor.

\section{Describing the NDS model}
\label{sec:nds_model}
In\cite{Crook2005}, Crook et al. have proposed a chaotic spiking neuron model that is called the NDS neuron. The NDS neuron is a conceptual discretized model that is based on R\"{o}ssler's chaotic system \cite{Rossler1976}. The NDS model is a modified version of R\"{o}ssler's equations as described by equations~\ref{eqn:rslr_x}-~\ref{eqn:rslr_z} in section~\ref{sec:rossler}.

By varying the system parameters such as period length ${\tau}$, connection time delays and initial conditions, large number of distinct orbits with varying periodicity may be stabilised.

\begin{figure}[ht]
\centerline{\includegraphics[scale=0.5]{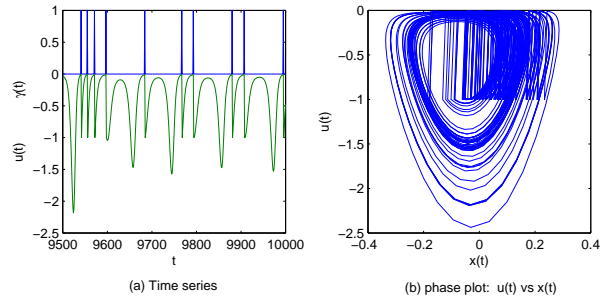}}
\caption{The chaotic behaviour of a NDS neuron without input (a) the time series of $u(t)$ and $\gamma(t)$, and (b) the phase space of $x(t)$ versus $u(t)$}
\label{fig:FIG03}
\end{figure}

The NDS model simulates a novel chaotic spiking neuron and is represented by:

\begin{equation}
\label{eqn:nds_x}
x(t+1) =x(t)+b(-y(t)-u(t))
\end{equation}

\begin{equation}
\label{eqn:nds_y}
y(t+1) = y(t)+c(x(t)+ay(t))
\end{equation}

\begin{equation}
\label{eqn:nds_u}
u(t+1) =
\left\{
\begin{array}{ll}
\eta\sb{0} & u(t)>\theta\\
u(t)+d(v+u(t)(-x(t))+ku(t))+F(t)+In(t) & u(t)\leq \theta
\end{array}
\right.
\end{equation}

\begin{equation}
\label{eqn:nds_Fb}
F(t) = \sum_{j=1}^{n} w_j \gamma(t-\tau_j)
\end{equation}

\begin{equation}
\label{eqn:nds_In}
In(t) = \sum_{j=1}^{n}I_j(t)
\end{equation}

\begin{equation}
\label{eqn:nds_gamma}
\gamma(t+1) =
\left\{
\begin{array}{ll}
1 & u(t)>\theta\\
0 & u(t)\leq \theta
\end{array}
\right.
\end{equation}

where ${x(t), y(t)}$ and  ${u(t)}$ describe the internal state of the neuron, ${\gamma(t)}$  is
the neuron's binary output, $F(t)$ represent the feedback signals, $In(t)$ is the external input binary spike train, and the constants and parameters of the model are: ${a=v=0.002}$, ${b = c = 0.03}$, ${d = 0.8}$, ${k = -0.057}$, ${\theta = -0.01}$, ${\eta\sb{0}=-0.7}$ and $\tau\sb{j}$ is the period length of the feedback signals.

\begin{figure}[!ht]
\centerline{\includegraphics[scale=0.5]{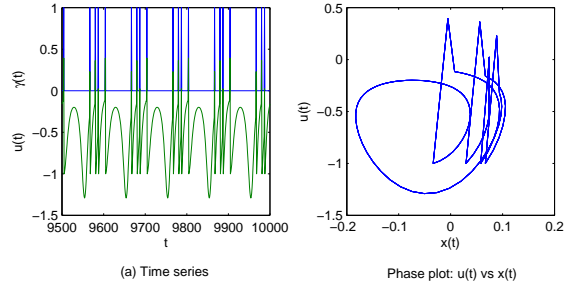}}
\caption{The stabilizing of period-4 orbit of a NDS neuron with feedback connection.} \label{fig:FIG04}
\end{figure}

The NDS model is a discrete version of the R\"{o}ssler system. The main reason to have a discrete version of the R\"{o}ssler system is because spikes should occur in discrete time. The discretization has been carried out by scaling the system variables ${x(t), y(t)}$ and  ${u(t)}$ using different scaling constants: $b,c,d$. The values of these constants have been tuned experimentally until the dynamics of the R\"{o}ssler system are preserved. If the values of these constants are large, then a system trajectory will miss many dynamic evolutions while moving from one discrete iteration to the next. Therefore, the time steps of the discrete system need to be very small so that a system trajectory will involve most of the dynamic evolutions of the R\"{o}ssler system. This will become clear in section~\ref{sec:discrete}.

The dynamics of a single NDS neuron without input is shown in Figure~\ref{fig:FIG03}. However, when the NDS neuron is equipped with a time delayed feedback connection then the firing pattern of the neuron can become periodic. This is shown in figure~\ref{fig:FIG04} where period-4 orbit is stabilised due to the feedback control mechanism $F$.

\section{Discretization method}
\label{sec:discrete}
There are two kinds of discretization methods when converting a continuous system into a discrete one: standard and nonstandard methods. One of the nonstandard methods is Euler’s Forward differentiation method that is usually used in developing simple simulation systems. To convert a continuous system into the corresponding discrete one using this method, then a time step $TS$ is used to approximate the next value of a continuous system variable that it will evolve to. For example, equation\ref{eqn:rslr_x} when converted into a discrete equation using Euler’s Forward differentiation it will become:

\begin{equation}
\label{eqn:rslr_x_dis1} \dot{x}(t_{k}) = \frac{x(t_{k+1})-x(t_{k})}{TS} = -y(t_{k})-u(t_{k})
\end{equation}

and then solving for $x(t_{k+1})$ gives:

\begin{equation}
\label{eqn:rslr_x_dis2} x(t_{k+1}) = x(t_{k})+TS(-y(t_{k})-u(t_{k}))
\end{equation}

After using the simple notation $x(t)$ instead of $x(t_{k})$, the equation becomes:

\begin{equation}
\label{eqn:rslr_x_dis3} x(t+1) = x(t)+TS(-y(t)-u(t))
\end{equation}

if this equation is compared with equation \ref{eqn:nds_x}, it is obvious that the $TS$ is chosen to be parameter $b=0.03$.

In order for the discretized function to behave similarly to the continuous one then $TS$ should be chosen to be small. For simulation purposes it is preferable to choose $TS$ according to the following formula:

\begin{equation}
\label{eqn:ts} TS \leq \frac{0.1}{|\lambda|_{max}}
\end{equation}

Where $|\lambda|_{max}$ is the largest absolute eigenvalues for the R\"{o}ssler system. According to the mathematical analysis that is carried out in \cite{Alhawarat2007}, $|\lambda|_{max}=5.68698$. When substituting this value in equation \ref{eqn:ts} it becomes:

\begin{equation}
\label{eqn:ts} TS \leq {0.0176}
\end{equation}

If this compared with $b$ it is obvious that the time step that has been chosen doesn't follow the simulation preferable setup.

Moreover, if we look at equation \ref{eqn:nds_y} not only the $TS$ is chosen to be $c=0.03$ but also a scaling factor is used to reduce the $y$ down by the factor $a=0.002$.

For the third variable in R\"{o}ssler system, viz. $z$ variable, there were many changes because the authors of the model according to \cite{Crook2005} wanted to invent a spiking model that is based on a threshold variable. That variable was $u$ which corresponds to $z$ in the original R\"{o}ssler system. In addition, the time step $TS$ value that is used in the discretization process was different from those appear in equations \ref{eqn:nds_x}-\ref{eqn:nds_y}. While the $TS=b=c=0.03$ used in the aforementioned equations, $TS$ is chosen to be $TS=d=0.8$ in discretization of the variable $u$. Moreover, the authors have scaled the constant $c$ from $5.7$ down to $0.057$ and changed its sign to negative. A final change was made to the sign of $x$ variable from positive to negative.

All these changes made the new system to behave differently in phase space. To summarize, the new attractor of the system has different fixed points types. According to the mathematical analysis carried out by \cite{Alhawarat2007}, the original R\"{o}ssler system fixed points, which are two spiral saddle points, have become two spiral repellors due to the varying scaling factors used and the change of the sign for both $k$ and $x$.

These results assure that the NDS model, although has a promising results as a spiking chaotic neuron model, it doesn't have a strong connection to the properties of the original R\"{o}ssler attractor. This is made obvious in \cite{Alhawarat2007} when they concluded that the existence of the UPOs of the NDS attractor
is due to the acting forces of the two spiral repellors and the reset mechanism. Because  without the reset mechanism, the two spiral repellors will enforce any trajectory that starts near by to evolve away from both of them and approaches infinity.

\begin{figure}
\centerline{\includegraphics[scale=0.8]{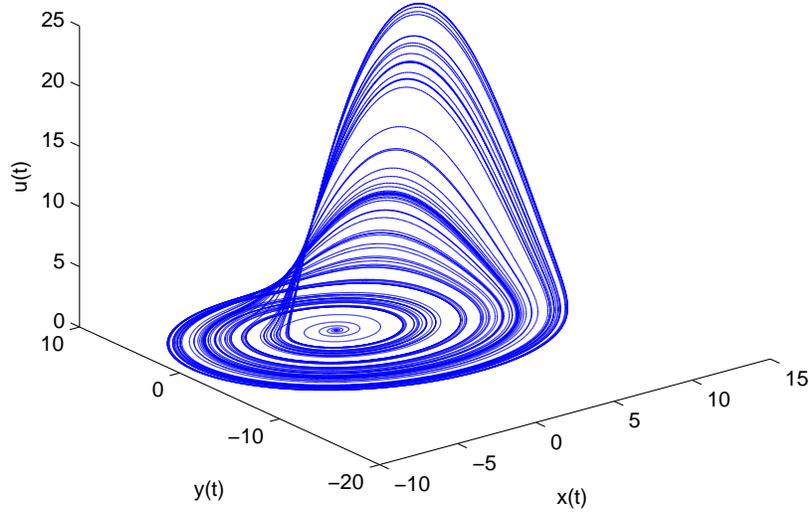}}
\caption{The Discrete version of R\"{o}ssler attractor based on equations~\ref{eqn:new_ross_x}-~\ref{eqn:new_ross_u}}
\label{fig:FIG05}
\end{figure}

If the Euler’s Forward differentiation discretization method is used to convert the continuous R\"{o}ssler system into a discrete model, where the time step is chosen to be $TS=0.0055$, then equations~\ref{eqn:rslr_x}-~\ref{eqn:rslr_z} will become:

\begin{equation}
\label{eqn:new_ross_x}
x(t+1) =x(t)+TS(-y(t)-u(t))
\end{equation}

\begin{equation}
\label{eqn:new_ross_y}
y(t+1) = y(t)+TS(x(t)+ay(t))
\end{equation}

\begin{equation}
\label{eqn:new_ross_u}
u(t+1) =u(t)+TS(b+z(t)(x(t)-c))
\end{equation}

Where ${TS = 0.0055}$, ${a=0.2}$, $b=0.2$, and ${c = 5.7}$.

To verify if such settings will preserve the shape and properties of the original R\"{o}ssler system, then an experiment is carried out to depict the discretized system's attractor as shown in figure~\ref{fig:FIG05}.

If these equations are converted into the equations of the NDS model then different steps need to be followed:

\begin{enumerate}
  \item Specify the value of $TS$, here $TS$ is chosen to be $0.0055$
  \item Change the parameter $b$ in equation~\ref{eqn:new_ross_u} to $v$.
  \item Change the parameter $c$ in equation~\ref{eqn:new_ross_u} to $k$.
  \item Change the constant $TS$ in equation~\ref{eqn:new_ross_x} to $b$ and give it the value of $TS$.
  \item Change the constant $TS$ in equation~\ref{eqn:new_ross_y} to $c$ and give it the value of $TS$.
  \item Change the constant $TS$ in equation~\ref{eqn:new_ross_u} to $d$ and give it the value of $TS$.
  \item Change the value of $a$ and $v$ and give it the value of $0.2$.
  \item Change the value of $k$ to $5.7$
  \item Change the variable $z$ to $u$.
  \item Change the sign of the term  $(x(t)-k)$ in equation~\ref{eqn:new_ross_u} to become $(-x(t)+k)$
\end{enumerate}

After applying the previous changes the equations~\ref{eqn:new_ross_x}-~\ref{eqn:new_ross_u} become:

\begin{equation}
\label{eqn:new_nds_x}
x(t+1) =x(t)+b(-y(t)-u(t))
\end{equation}

\begin{equation}
\label{eqn:new_nds_y}
y(t+1) = y(t)+c(x(t)+ay(t))
\end{equation}

\begin{equation}
\label{eqn:new_nds_u}
u(t+1) =u(t)+d(v+u(t)(-x(t)+k))
\end{equation}

Where ${a=0.2}$, $b=c=d=0.0055$, and ${k=5.7}$.

\begin{figure}[!ht]
\centerline{\includegraphics[scale=0.8]{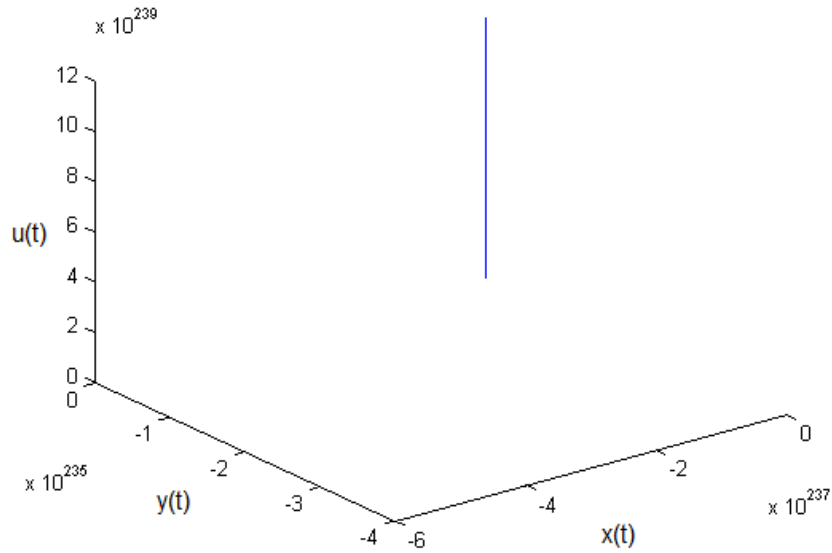}}
\caption{The attractor of the model that is based on equations~\ref{eqn:new_nds_x}-~\ref{eqn:new_nds_u}}
\label{fig:FIG06}
\end{figure}

Now this new model need to be verified, i.e. an experiment need to be carried out to depict the attractor such equations will produce. Figure~\ref{fig:FIG06} depicts the results of evolution of equations~\ref{eqn:new_nds_x}-~\ref{eqn:new_nds_u}.

\FloatBarrier

Note that the original R\"{o}ssler attractor has disappeared and this is due to the change of sign that is made to the term $(x(t)-k)$ in equation~\ref{eqn:new_nds_u}.

\section{Tuning the parameters of the NDS model}
\label{sec:tuning}
In this research, many experiments have been carried out to tune the parameters of the NDS model. These experiments are carried out by varying one parameter and fix all other parameters to their default values that are used in the NDS model definition as stated in section \ref{sec:nds_model}. The following parameters have been considered in these experiments: $a,v,b,c,d$ and $k$ because the other parameters: $\theta$, $\eta\sb{0}$ and $\tau\sb{j}$ have already been studied thoroughly in \cite{Alhawarat2007}. One NDS neuron has been used in the experiments setup and random initial conditions are chosen for the values of the variables: $x,y$ and $u$. After 1000 iterations the feedback control is applied and the experiment runs for another 9000 iterations.

To decide whether a specific setting is valid; the values of the variables of the model are recorded and then depicted in phase space. If an attractor exist, then this setting is considered in the valid ranges for the values of the system parameters.

\begin{table}[!ht]
\caption{Parameter value's ranges}
\label{tab:results}
\begin{tabular}{lccccc}
\hline\noalign{\smallskip}
Parameter & \,\,\,\, $a,v$ & \,\,\,\, $b,c$ & \,\,\,\, $d$ & \,\,\,\, $k$ \\
\noalign{\smallskip}
\hline
\noalign{\smallskip}
  Range & \,\,\,\, $0.001-0.1$ & \,\,\,\, $0.01-0.055$ & \,\,\,\, $0.8-0.9$ & \,\,\,\, $-(0.055-0.58)$ \\
\hline
\end{tabular}
\end{table}

To summarize the results; the valid ranges for the model parameters are shown in table~\ref{tab:results}.

\FloatBarrier
\begin{table}[!ht]
\caption{Parameter settings with different selected values from the ranges appear in table\ref{tab:results}}
\label{tab:param_settings}
\begin{tabular}{cccccc}
\hline\noalign{\smallskip}

Parameter & \,\,\,\, $a,v$ & \,\,\,\, $b,c$ & \,\,\,\, $d$ & \,\,\,\, $k$ \\
\noalign{\smallskip}
\hline
\noalign{\smallskip}
  Setup 01 & \,\,\,\, $0.001$ & \,\,\,\, $0.03$ & \,\,\,\, $0.8$ & \,\,\,\, $-0.057$ \\
  Setup 02 & \,\,\,\, $0.01$ & \,\,\,\, $0.03$ & \,\,\,\, $0.8$ & \,\,\,\, $-0.057$ \\
  Setup 03 & \,\,\,\, $0.1$ & \,\,\,\, $0.03$ & \,\,\,\, $0.8$ & \,\,\,\, $-0.057$ \\
  Setup 04 & \,\,\,\, $0.002$ & \,\,\,\, $0.001$ & \,\,\,\, $0.8$ & \,\,\,\, $-0.057$ \\
  Setup 05 & \,\,\,\, $0.002$ & \,\,\,\, $0.02$ & \,\,\,\, $0.8$ & \,\,\,\, $-0.057$ \\
  Setup 06 & \,\,\,\, $0.002$ & \,\,\,\, $0.05$ & \,\,\,\, $0.8$ & \,\,\,\, $-0.057$ \\
  Setup 07 & \,\,\,\, $0.002$ & \,\,\,\, $0.03$ & \,\,\,\, $0.8$ & \,\,\,\, $-0.057$ \\
  Setup 08 & \,\,\,\, $0.002$ & \,\,\,\, $0.03$ & \,\,\,\, $0.85$ & \,\,\,\, $-0.057$ \\
  Setup 09 & \,\,\,\, $0.002$ & \,\,\,\, $0.03$ & \,\,\,\, $0.9$ & \,\,\,\, $-0.057$ \\
  Setup 10 & \,\,\,\, $0.002$ & \,\,\,\, $0.03$ & \,\,\,\, $0.8$ & \,\,\,\, $-0.055$ \\
  Setup 11 & \,\,\,\, $0.002$ & \,\,\,\, $0.03$ & \,\,\,\, $0.8$ & \,\,\,\, $-0.056$ \\
  Setup 12 & \,\,\,\, $0.002$ & \,\,\,\, $0.03$ & \,\,\,\, $0.8$ & \,\,\,\, $-0.058$ \\
  Setup 13 & \,\,\,\, $0.01$ & \,\,\,\, $0.05$ & \,\,\,\, $0.85$ & \,\,\,\, $-0.055$ \\
  Setup 14 & \,\,\,\, $0.002$ & \,\,\,\, $0.015$ & \,\,\,\, $0.8$ & \,\,\,\, $-0.058$ \\
  Setup 15 & \,\,\,\, $0.1$ & \,\,\,\, $0.04$ & \,\,\,\, $0.8$ & \,\,\,\, $-0.056$ \\

\hline
\end{tabular}
\end{table}

\FloatBarrier
\begin{figure}[!ht]
\centerline{\includegraphics[scale=0.42]{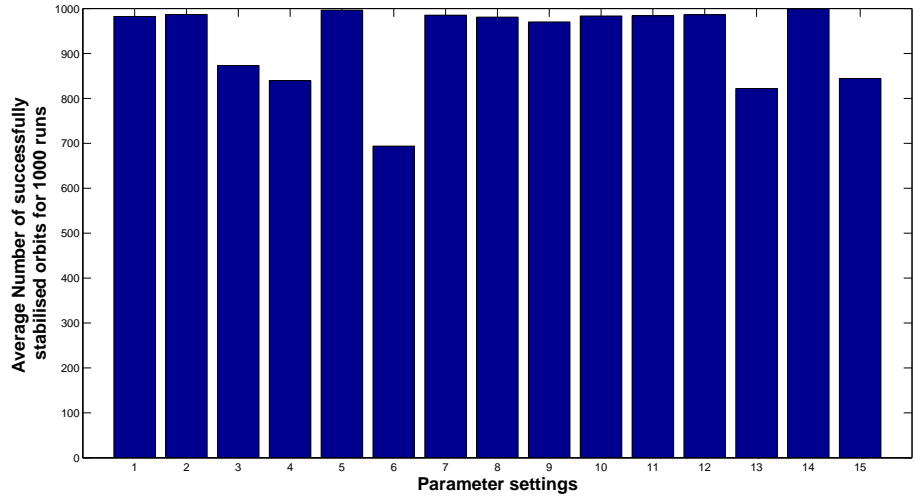}}
\caption{Average Stabilized UPOs over 1000 run based on the parameter settings that appear on table~\ref{tab:param_settings}.} \label{fig:FIG07}
\end{figure}

To judge weather such ranges could enhance the capacity of the attractor in terms of the number of UPOs that it might encompass, then another experiment setup is used. Here the average number of successfully stabilized UPOs is computed over 1000 run for different parameter settings according to table~\ref{tab:param_settings} and the results are depicted in figure~\ref{fig:FIG06}.

Note that the original parameters values of the NDS model is used in setup 7. This is done to compare the
capacity of the attractor with different parameter settings compared to the original parameter settings.

The results that are shown in figure~\ref{fig:FIG06} suggest that there exist better parameter settings such as Settings $14$ and Settings $05$, but these will not make a significant enhancement in the capacity of the attractor in terms of successfully stabilized UPOs when compared to the NDS original settings (Setting 07).

Based on the results that have been attained in section~\ref{sec:discrete}, if a parameter setting need to be chosen to represent the R\"{o}ssler system then first a $TS$ need to be set according to equation~\ref{eqn:ts}, then a mapping between the original R\"{o}ssler equations and the NDS equations need to be performed. If $TS$ is chosen to be $0.015$, then after carrying out the mapping between the equations of both systems; the parameter settings become as listed in table~\ref{tab:ross}

\FloatBarrier
\begin{table}[!ht]
\caption{Parameter settings for the original R\"{o}ssler as discussed in section~\ref{sec:rossler}}
\label{tab:ross}
\begin{tabular}{lccccc}
\hline\noalign{\smallskip}
Parameter & \,\,\,\, $a,v$ & \,\,\,\, $b,c$ & \,\,\,\, $d$ & \,\,\,\, $k$ \\
\noalign{\smallskip}
\hline
\noalign{\smallskip}
  Parameter value & \,\,\,\, $0.2$ & \,\,\,\, $0.015$ & \,\,\,\, $0.015$ & \,\,\,\, $5.7$ \\
\hline
\end{tabular}
\end{table}

This setting is considered in another experiment setup, and the result is that there is no single UPO that can be stabilized. One possible reason is that the fixed points of the system is reserved in this case and not changed as the case in the NDS model settings. The change in sign in the term $(x-c)$ and the different nested scaling factors that are applied to the original R\"{o}ssler equation have led to change in the types and properties of the fixed points as shown in~\cite{Alhawarat2007}.

This can be proved easily because when multiplying an equation with a constant and then trying to find its roots, then it should be set to zero. Therefore, the constant will have no effect on the resulted roots as dividing both sides by constant will eliminate the constant from the left hand side and will not affect zero in the right hand side.
\section{Discussion}
\label{sec:discussion}
The results of both experimentations and mathematical analysis of the discretization method suggest that the NDS model has weak connections to the original R\"{o}ssler system. This is due to many factors. Firstly, the discretization process that is used in devising the NDS equations does not follow any known discretization method where different discretization time steps and scaling factors are used. Secondly, Changing the sign of the term $(x(t)-k)$ that appears in equation~\ref{eqn:new_nds_u} made the system attractor to disappear and approach infinity instead. These changes affected both the shape and properties of the original R\"{o}ssler attractor.

It is important here to stress that even before adding the feedback signal, the input signals and the reset mechanism to the system equations, the attractor has become completely different from the original R\"{o}ssler attractor. This assure that the NDS model has weak connections to the R\"{o}ssler model and has different fixed points, eigenvalues and eigenvectors as demonstrated in~\cite{Alhawarat2007}.

The results also suggest that both the reset mechanism and the feedback signal are the major ingredients for the NDS model to work and to be stabilised to one of its available UPOs.

The results of experimental setups made to tune the NDS parameters suggest that there exist better parameter settings but will not enhance the capacity of the attractor in terms of successfully stabilized UPOs significantly when compared to the NDS original settings.

Also, the settings of the R\"{o}ssler model when used in the NDS model resulted in no stabilized UPOs because the discretization method that is used to build the NDS model has led to changes in the shape and properties of the R\"{o}ssler attractor.

It is important to mention that the main factors that affect the shape and properties of the original R\"{o}ssler attractor are both scaling the parameters of the model with different values and the change in sign that is made to the term $(x(t)-k)$ that appears in equation~\ref{eqn:new_nds_u}.

\section{Conclusion}
\label{sec:conclusion}
One chaotic model, viz., the NDS model has been studied in this paper. NDS is one of different chaotic models that are devised in recent years to explore the possibilities of exploiting the rich dynamics that such models might provide for carrying out information processing tasks.

The NDS model might be stabilized to a large number of UPOs. These UPOs can be stabilised using a feedback control mechanism. The NDS model is a modified version of R\"{o}ssler chaotic system. The rich dynamics of the R\"{o}ssler system is supposed to be inherited by the NDS model. This is suggested by the large number of UPOs that can be stabilised as shown in figure~\ref{fig:FIG03}.

However, when the discretization methods are discussed in this paper, it is shown that the method that is used to discretize the original R\"{o}ssler equations in devising the NDS equations is not a known discretization method. This along with the change in sign in the term $u(t)x(t)-ku(t))$ has affected the shape and properties of the NDS attractor when compared with its origin: the R\"{o}ssler attractor.

Different experimental setups have been prepared and performed to tune the NDS model parameters. The results of these experimentations have revealed the valid ranges of the parameters of the model. Also, other experimentations have shown different capacities for the NDS attractor in terms of the number of stabilized UPOs with different parameter settings.

The results attained in this paper suggest that there are weak relationships between the NDS and the R\"{o}ssler models. However, the NDS attractor encompasses large number of UPOs as shown in figure~\ref{fig:FIG07}. These and the wide range of dynamic behaviours may be exploited to carry out information processing tasks.

\begin{acknowledgement}
This work was supported by Salman Bin Abdulaziz University in KSA under grant 40/H/1432. I'm grateful to this support and would like to thank the deanship of scientific research at the university, represented by Dr. Abdullah Alqahtani at that time.
\end{acknowledgement}

% BibTeX users please use one of %\bibliographystyle{spbasic}      % basic style, author-year citations
\bibliographystyle{spmpsci}
\bibliography{MohdPapers}   % name your BibTeX data base

\end{document}